\documentclass[letterpaper]{article} 
\usepackage[table,xcdraw]{xcolor}
\usepackage{aaai2026}  
\usepackage{times}  
\usepackage{helvet}  
\usepackage{courier}  
\usepackage[hyphens]{url}  
\usepackage{graphicx} 
\usepackage{natbib}  
\usepackage{caption} 

\usepackage{booktabs}
\usepackage{colortbl} 
\usepackage{multirow}
\usepackage{pifont}
\usepackage{siunitx}
\usepackage{tabularx}
\usepackage{ragged2e}
\usepackage{float}
\definecolor{lightgrey}{HTML}{e0ffff}
\definecolor{lightgreen}{RGB}{127,221,250}
\usepackage{algorithm}
\usepackage{algpseudocode}
\usepackage{newfloat}
\usepackage{listings}
\usepackage{amsmath}
\usepackage[most]{tcolorbox} 

\usepackage[colorlinks=true, allcolors=darkblue]{hyperref}
\DeclareCaptionStyle{ruled}{labelfont=normalfont,labelsep=colon,strut=off} 
\floatstyle{ruled}
\newfloat{listing}{tb}{lst}{}
\floatname{listing}{Listing}

\usepackage{soul}     
\definecolor{lightblue}{RGB}{221, 231, 245} 
\definecolor{lightyellow}{RGB}{209, 239, 241}
\definecolor{lightgreen}{RGB}{255, 240, 230}
\definecolor{lightred}{RGB}{255,102,102}
\makeatletter
\newcommand*\myfontsize{%
  \@setfontsize\myfontsize{7}{8}%
  \rmfamily 
}
\makeatother
\usepackage{tikz}
\usetikzlibrary{tikzmark}
\newcommand{\mytextbox}[2]{%
  \,\raisebox{0.17ex}{\tikzmarknode[draw=#1, thick, inner sep=2.2pt]{#2}{\myfontsize \textbf{\textcolor{#1}{#2}}}}\,%
}
\definecolor{tryanswercolor}{HTML}{009933}
\definecolor{memcolor}{HTML}{333399}
\definecolor{memfusecolor}{HTML}{660099}
\definecolor{selfprobecolor}{HTML}{993300}
\definecolor{darkblue}{rgb}{0, 0, 0.5}

\newcommand{\tret}{\mytextbox{black}{Tri-Retrieve}}
\newcommand{\tans}{\mytextbox{tryanswercolor}{Try-Answer}}
\newcommand{\menc}{\mytextbox{memcolor}{Mem-Encode}}
\newcommand{\mfuse}{\mytextbox{memfusecolor}{Mem-Fuse}}
\newcommand{\mupdate}{\mytextbox{memcolor}{Mem-Update}}
\newcommand{\smon}{\mytextbox{selfprobecolor}{Self-Probe}}

\newcommand{\tansw}{\mytextbox{white}{Try-Answer}}
\newcommand{\mencw}{\mytextbox{white}{Mem-Encode}}
\newcommand{\mfusew}{\mytextbox{white}{Mem-Fuse}}

\newcommand{\smonw}{\mytextbox{white}{Self-Probe}}
\usepackage{xspace}
\newcommand{\rag}{ComoRAG\xspace}
\usepackage{enumitem}

\title{ComoRAG: A \underline{Co}gnitive-Inspired \underline{M}emory-\underline{O}rganized RAG\\for Stateful Long Narrative Reasoning}

\author{
    Juyuan Wang\textsuperscript{\rm 1}\equalcontrib~~
    Rongchen Zhao\textsuperscript{\rm 1}\equalcontrib~~ 
    Wei Wei\textsuperscript{\rm 2}~~
    Yufeng Wang\textsuperscript{\rm 1}\\ 
    Mo Yu\textsuperscript{\rm 4}~~~
    Jie Zhou\textsuperscript{\rm 4}~~~
    Jin Xu\textsuperscript{\rm 1,3}~~~
    Liyan Xu\textsuperscript{\rm 4}\thanks{Project lead. Correspondence to: \textless \href{mailto:liyanlxu@tencent.com}{liyanlxu@tencent.com}\textgreater}
}
\affiliations{
    \textsuperscript{\rm 1}School of Future Technology, South China University of Technology\\
    \textsuperscript{\rm 2}Independent Researcher~~~~~
    \textsuperscript{\rm 3}Pazhou Lab, Guangzhou\\
    \textsuperscript{\rm 4}WeChat AI, Tencent
}

\usepackage{bibentry}

\begin{document}

\maketitle

\begin{abstract}
Narrative comprehension on long stories and novels has been a challenging domain attributed to their intricate plotlines and entangled, often evolving relations among characters and entities.
Given the LLM's diminished reasoning over extended context and its high computational cost, retrieval-based approaches remain a pivotal role in practice.
However, traditional RAG methods could fall short due to their stateless, single-step retrieval process, which often overlooks the dynamic nature of capturing interconnected relations within long-range context.
In this work, we propose ComoRAG, holding the principle that narrative reasoning is not a one-shot process, but a dynamic, evolving interplay between new evidence acquisition and past knowledge consolidation, analogous to human cognition on reasoning with memory-related signals in the brain.
Specifically, when encountering a reasoning impasse, ComoRAG undergoes iterative reasoning cycles while interacting with a dynamic memory workspace. In each cycle, it generates probing queries to devise new exploratory paths, then integrates the retrieved evidence of new aspects into a global memory pool, thereby supporting the emergence of a coherent context for the query resolution.
Across four challenging long-context narrative benchmarks (200K+ tokens), ComoRAG outperforms strong RAG baselines with consistent relative gains up to 11\% compared to the strongest baseline. Further analysis reveals that ComoRAG is particularly advantageous for complex queries requiring global context comprehension, offering a principled, cognitively motivated paradigm towards retrieval-based stateful reasoning. Our framework is made publicly available at \url{https://github.com/EternityJune25/ComoRAG}.
\end{abstract}

\section{Introduction}

The core challenge of long narrative comprehension lies not merely in connecting discrete pieces of evidence, a task more naturally defined as multi-hop Question Answering (QA), but in performing a \textbf{dynamic cognitive synthesis} to grasp necessary background and content progression \cite{xu-etal-2024-fine}. Unlike multi-hop QA \cite{yang2018hotpotqa}, which seeks a static path through fixed facts, narrative comprehension requires emulating a human reader: continuously building and revising a \textbf{global mental model} of the plot, characters, and their evolving motivations \cite{johnson1983mental}. The complexity of this process is well exemplified by a classic question \textit{``Why did Snape kill Dumbledore?''} from the Harry Potter series. Answering this requires weaving a complete web of evidence from disparate clues spanning multiple books—Dumbledore's terminal illness, the Unbreakable Vow, and Snape's deeply concealed loyalty. The true significance of these clues is only fully reconciled in hindsight. This capability is what we term \textbf{stateful reasoning}: it demands more than linking static evidence; it requires maintaining a \emph{dynamic memory} of the narrative, one that is constantly updated as new revelations emerge.
\begin{figure}[t]
\centering
\includegraphics[width=0.9\columnwidth]{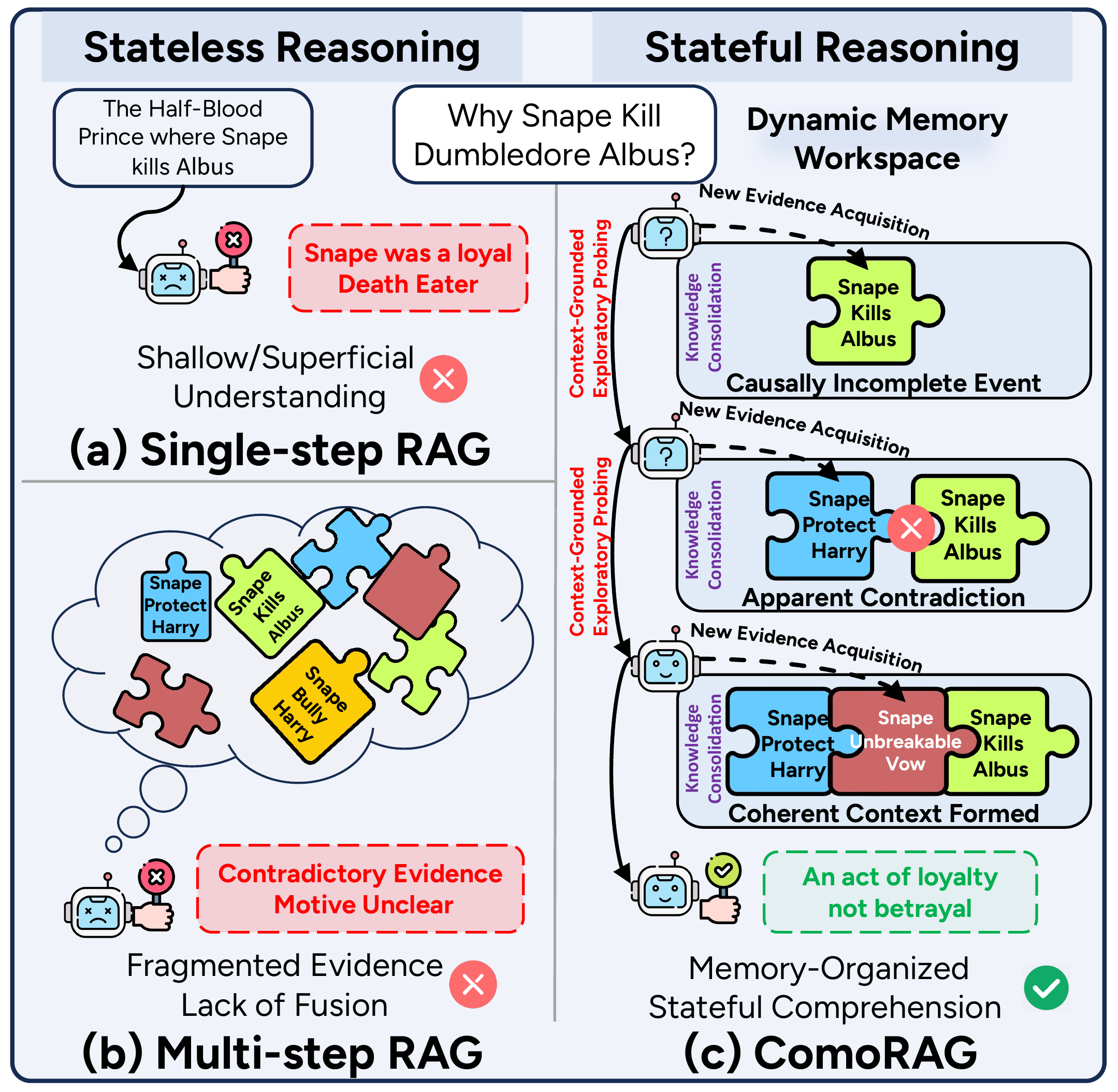}
\caption{Comparison of RAG reasoning paradigms.}
\label{fig1}
\vspace{-1.5ex}
\end{figure}
Long-context LLMs have demonstrated promising performance on benchmarks such as the ``Needle in a Haystack'' \cite{needle}. However, their capacity to process long narratives (200k+ tokens)  remains limited by finite context windows. Furthermore, as the input length increases, these models are prone to the ``lost in the middle'' problem \cite{lost_in_the_Middle}, which raises perplexity and impairs generation quality. This limitation is particularly pronounced in narrative tasks which require stateful reasoning. As a result, retrieval-augmented generation (RAG) \cite{NEURIPS2020_6b493230} has emerged as an important direction for tackling long context comprehension with LLMs, leveraging text embeddings or more advanced retrieval paradigms such as embeddings situated on global context \cite{sitemb}.

However, existing RAG methods still struggle to effectively address this challenge. Advanced single-step retrieval remains limited by its static index. This includes methods such as RAPTOR \cite{raptor}, which clusters and summarizes text chunks to retrieve at different levels of details; HippoRAGv2 \cite{hipporag2} and GraphRAG \cite{graphrag}, which build knowledge graphs to achieve multi-hop reasoning in a single retrieval step. Nonetheless, one-shot static retrieval inevitably leads to shallow comprehension. For example, the evidence about Snape in Fig.~\ref{fig1}(a) can mislead the model into making a false inference.

As a remedy, multi-step retrieval methods offer a more promising direction, such as IRCoT \cite{ircot}, which interleaves the retrieval process with Chain-of-Thought reasoning \cite{cot}; Self-RAG \cite{selfrag}, which trains a model to adaptively retrieve and reflect on evidence; and MemoRAG \cite{memorag}, which uses a dual-system architecture to generate clues from compressed global context. These methods all target to obtain richer context through iterative retrieval. However, their retrieval steps are typically independent, which lack coherent reasoning throughout explicit narrative progression, featuring fragmented evidence with a stateless comprehension. 
As illustrated in Figure~\ref{fig1}(b), due to a lack of dynamic memory, multi-step retrieval fails to integrate contradictory evidence such as \textit{``Snape protects/bullies Harry''} and cannot understand the evolution of his actions, ultimately unable to yield the correct answer.

In this work, we seek inspiration from the function of Prefrontal Cortex (PFC) in human brains, which employs a sophisticated reasoning process called \textbf{Metacognitive Regulation} \cite{FERNANDEZDUQUE2000288}. This process is not a single action but a dynamic interplay between \textbf{new evidence acquisition}, driven by goal-directed \textbf{memory probes} \cite{10.1162/jocn.2006.18.9.1439,miller2024timescales}, and subsequent \textbf{knowledge consolidation}. During consolidation, new findings are integrated with past information to construct an evolving, coherent narrative. This iterative cycle allows the PFC to continuously assess its understanding and revise its strategy, providing a direct cognitive blueprint for our framework's stateful reasoning approach.

We introduce \rag, a \underline{co}gnitive-inspired, \underline{m}emory-\underline{o}rganized RAG framework, imitating the human Prefrontal Cortex (PFC) for achieving stateful reasoning. At its core is a dynamic cognitive loop operating on a memory workspace, which actively probes and integrates new evidence to build a coherent narrative comprehension.

This process, as illustrated in Figure~\ref{fig1}(c), is a closed loop of evolving reasoning states. Faced with a complex query like \textit{``Why did Snape kill Dumbledore?''}, the system's memory state evolves from an initial ``causally incomplete event'' (\emph{Snape kills Albus}), to an ``apparent contradiction'' upon finding contradictory information (\emph{Snape protects Harry}), and ultimately to a logically consistent \textbf{coherent context} through deeper exploration and evidence fusion. Only in this final, complete cognitive state can \rag perform the correct stateful reasoning, deriving the true insight that it was \emph{``an act of loyalty, not betrayal''}.

This cognitively-inspired design yields substantial improvements across four challenging long-context narrative benchmarks. \rag is shown to consistently outperform all categories of strong baselines across each dataset. Our analysis reveals several key findings. First, these gains stem directly from the cognitive loop, which transforms a static knowledge base into a dynamic reasoning engine; for instance, accuracy on EN.MC jumps from a static-retrieval baseline of 64.6\% to 72.9\%, with performance efficiently converging in around 2-3 cycles. Second, our framework excels on \emph{narrative queries} that require global understanding of plot progression, achieving up to a \textbf{19\%} relative F1 improvement on these challenging question types where others falter. Finally, our framework demonstrates remarkable modularity and generalizability. Its core loop can be flexibly integrated to existing RAG methods such as RAPTOR, which directly yields a 21\% relative accuracy gain). Also, switching to a stronger model as the backbone LLM agents can upgrade reasoning in the entire cognitive loop, attaining accuracy from 72.93\% to 78.17\%. 
These results collectively validate that \rag provides a principled, cognitively-inspired new paradigm for retrieval-based long narrative comprehension towards stateful reasoning.

\begin{figure*}[t]
\centering
\includegraphics[width=1\textwidth]{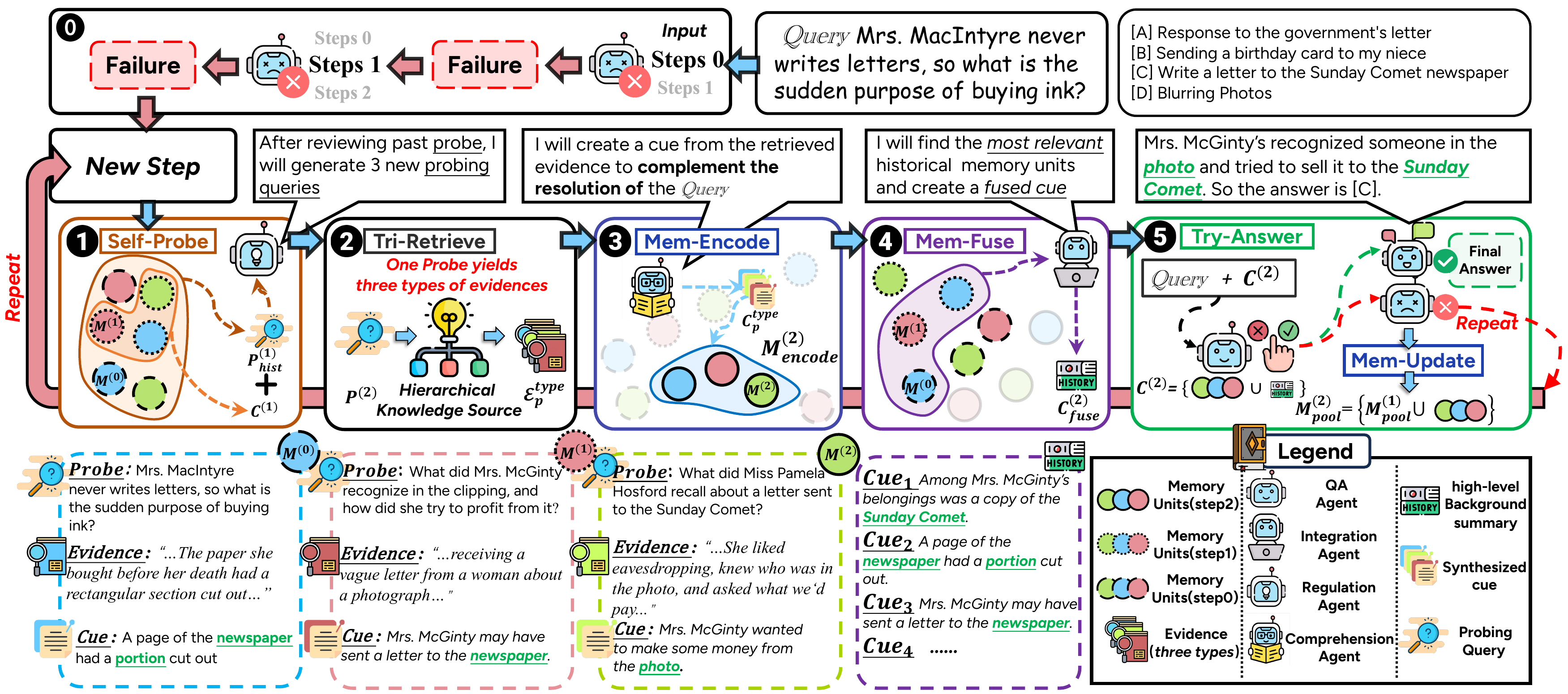} %
\caption{\textbf{An illustration of \rag.} Triggered by a reasoning impasse (Failure), the Metacognitive Regulation loop consists of five core operations described in Section~\ref{ssec:loop}: 1) \textbf{\smon} to devise new exploratory probing queries based on past memory units; 2) \textbf{\tret} to retrieve evidence from three knowledge sources; 3) \textbf{\menc} to form new memory units on how the latest evidence of new aspects could complement the final query resolution; 4) \textbf{\mfuse} to generate cues integrating new and past memory units; 5) \textbf{\tans} to perform query answering using new memory information produced in this cycle.}
\label{fig:overview_comorag}
\end{figure*}

\section{Methodology}
\label{sec:method}

We introduce ComoRAG, an autonomous cognitive architecture designed to formalize and implement the process of \textbf{Metacognitive Regulation} outlined in the Introduction. The architecture's design is directly inspired by the functional mechanisms of the Prefrontal Cortex (PFC) and is founded on three conceptual pillars: (1) a \textbf{Hierarchical Knowledge Source} for deep contextual understanding; (2) a \textbf{Dynamic Memory Workspace} for tracking and integrating the multi-turn reasoning; and (3) a \textbf{Metacognitive Control Loop} that drives the entire resolving procedure. 

\subsection{Problem Formulation: Towards Principled Narrative Reasoning}
Our objective is to design a framework for stateful reasoning in RAG scenarios. Especially, it aims to resolve those queries that require global context comprehension in the first place, commonly seen in narratives, where conventional RAG may fail to recognize relevant context based on the surface form of queries.
Formally, denote the initial query as $q_{init}$, and a knowledge source $\mathcal{X}$ derived upon the original context, our framework $F$ leverages a series of adaptive operations to yield the final answer, $A_{final}$, through discrete time steps $t=1, \dots, T$ with underlying memory control.

At the beginning of each step $t$, $F$ determines its focus of reasoning—a set of new probing queries $\mathcal{P}^{(t)}$, representing new information to seek that may logically deepen the query comprehension and ultimately complement the answer resolution. 
With newly retrieved information by $\mathcal{P}^{(t)}$ at each step, the framework utilizes the global memory pool maintained till the prior step $\mathcal{M}_{pool}^{(t-1)}$, and produces either the final answer, or a Failure Signal, indicating a reasoning impasse—and updates the memory pool to $\mathcal{M}_{pool}^{(t)}$, accomplishing a cognitive cycle that synergizes between the knowledge source, memory space and retrieval operations.

\subsection{The Hierarchical Knowledge Source}
\label{ssec:index}

To overcome the limitations of a monolithic representation of the given context, our framework first builds a hierarchical knowledge index $\mathcal{X}$ for retrieval that models the raw text from three complementary cognitive dimensions, analogous to how the PFC integrates different memory types from various brain regions, particularly supporting cross-layer reasoning from raw evidence to abstract relationships.

\paragraph{Veridical Layer: Grounding in Factual Evidence.}
To ensure all reasoning is traceable to source evidence, a veridical layer $\mathcal{X}^{ver}$ is firstly established, constituted by raw text chunks directly, analogous to the precise recall of factual details in human memory. For more accurate retrieval on text chunks, we instruct a LLM to generate knowledge triples (\emph{subject}-\emph{predicate}-\emph{object}) for each text chunk. These triples participate in each retrieval, and strengthen the matching between an incoming query and the corresponding text chunk, which is proven effective by HippoRAG \cite{hipporag1}. Further details are described in Appendix~\ref{sec:impl}.

\paragraph{Semantic Layer: Abstracting Thematic Structure.}
To capture thematic and conceptual connections that transcend across long-range contextual dependencies, a semantic layer $\mathcal{X}^{sem}$ is built, inspired by the prior work RAPTOR that employs a GMM-driven clustering algorithm to recursively summarize semantically similar text chunks into a hierarchical summary tree. We reckon such semantic abstraction is necessary for deeper comprehension and follow the same formulism. These summary nodes enable the framework to retrieve conceptual information beyond the surface level.

\paragraph{Episodic Layer: Reconstructing Narrative Flow.}
The previous two layers equip views of both factual details and high-level concepts. However, they lack temporal development or plot progression that can be especially crucial for narratives. To enable such view with long-range causal chains, we introduce the episodic layer, $\mathcal{X}^{epi}$, which aims to reconstruct the plotline and story arc by capturing the sequential narrative development.
The process features a sliding window summarization across text chunks; each resulting node is then a summary that aggregates the narrative development of continuous or causally related events according to the timeline. Optionally, the sliding window process can be applied recursively to form higher-level views of content progression, extracting different levels of narrative flow as part of the knowledge source.

\subsection{The Architecture of Metacognitive Regulation}
\label{ssec:loop}

The core of ComoRAG is a control loop that fully realizes the concept of metacognitive regulation. It is composed of a \textbf{Regulatory Process} for reflection and planning at each step, and a \textbf{Metacognitive Process} for executing reasoning and memory management with the \textbf{Memory Workspace}.

\paragraph{Dynamic Memory Workspace.}
The memory workspace contains memory units that serve as the bridge for a cohesive multi-step exploration and reasoning by metacognitive regulation.
Each memory unit $m$ functionally \textbf{concludes one retrieval operation}, denoted as a tuple of three elements: $m = (p, \mathcal{E}^{type}_p, \mathcal{C}^{type}_p)$, where $p$ is the \emph{probing query} that triggers this retrieval; $\mathcal{E}^{type}_p$ is the homogeneous set of evidence retrieved from a single knowledge layer ($type \in \{ver,sem,epi\}$); and $\mathcal{C}^{type}_p$ is a \emph{synthesized cue} that reflects how these retrieved evidence by the probe $p$ could complement the comprehension and resolution of the original query $q_{init}$.
Concretely, $\mathcal{C}^{type}_p$ is generated by a LLM in the role of Comprehension Agent, $\pi_{cue}$, denoted as $\mathcal{C}^{type}_p = \pi_{cue}(q_{init}, p, \mathcal{E}^{type}_p)$.

The formation of a memory unit $(p, \mathcal{E}^{type}_p, \mathcal{C}^{type}_p)$ by each retrieval is defined as a \textbf{\menc} operation. 
The memory workspace/pool will be utilized and updated throughout the reasoning cycle described below.

\paragraph{The Regulatory Process.}
The regulatory process is invoked at the beginning of a reasoning cycle/step $t$ if the preceding cycle $t-1$ is concluded in failure. The core operation, \textbf{\smon}, plans new probing queries of which retrieved information may contribute to the final answer, thereby devising new exploratory paths to break the impasse. It is orchestrated by a \textbf{Regulation Agent}, $\pi_{probe}$, whose decisions are informed by the reflection on the prior failure, exploring for more necessary background or relevant information towards a full context comprehension to resolve the original query.
\textbf{\smon} takes three inputs: (1) the ultimate goal $q_{init}$; (2) the complete exploration probing history $\mathcal{P}_{hist}^{(t-1)}$ up to the end of the last step; and (3) the immediate knowledge gaps that caused the failure, concretized by all \emph{synthesized cues} of memory units generated in the prior step, denoted as $\{\mathcal{C}\}^{(t-1)}$. Its output $\mathcal{P}^{(t)}$ is a new, strategic set of retrieving probes for the current cycle $t$:
\begin{equation}
\label{eq:smon}
\mathcal{P}^{(t)} = \pi_{probe}\big(q_{init},\;\mathcal{P}_{hist}^{(t-1)},\;\{\mathcal{C}\}^{(t-1)}\big)
\end{equation}

\paragraph{The Metacognitive Process.}
The metacognitive process takes the new probes for this cycle $\mathcal{P}^{(t)}$, and performs reasoning towards resolving the original query while keeping track of the progress with the memory space. It comprises a series of operations, described in details as follows.

\textbf{\tret}:
for each probing query $p \in \mathcal{P}^{(t)}$, a retrieval is conducted on each knowledge layer $\mathcal{X}^{type}$ where $type \in \{ver, sem, epi\}$, such that evidence of high embedding similarity to $p$ per layer is retrieved in a standard Dense Passage Retrieval paradigm, with each evidence being either the raw text chunk, a semantically clustered summary, or a narrative flow summary.

\textbf{\menc}:
for each $p$ and $type$, the retrieved evidence is immediately processed by the aforementioned \textbf{\menc}, to generate a new memory unit that keeps track of how this specific probing could complement to the final answer. The number of all generated memory units at this step can be denoted as $|\mathcal{M}^{(t)}_{encode}| = 3\times|\mathcal{P}^{(t)}|$. 

\textbf{\mfuse}:
new memory units in the above step $\mathcal{M}^{(t)}_{encode}$ mainly emphasize aspects probed in the current cycle. To fully utilize the past experience and historical knowledge, the framework further identifies relevant \emph{synthesized cues} from past units in the existing memory pool $\mathcal{M}^{{t-1}}_{pool}$, then generates a new synthesized cue for fusing past relevant evidence. Let $\mathcal{M}^{{t-1}}_{pool} \circ q_{init}$ represent past memory units whose cues are of high embedding similarity with $q_{init}$, and denote a LLM as \textbf{Integration Agent} $\pi_{fuse}$ that synthesizes these relevant past evidence into a high-level background summary, the new cue fusing past memory $\mathcal{C}_{fuse}^{(t)}$ is then:
\begin{equation}
\label{eq:fuse}
\mathcal{C}_{fuse}^{(t)} = \pi_{fuse} \big(q_{init},\;\mathcal{M}^{{t-1}}_{pool} \circ q_{init}\big)
\end{equation}

\textbf{\tans}:
with the new probing evidence in $\mathcal{M}^{(t)}_{encode}$ and the past-fusing cue $\mathcal{C}_{fuse}^{(t)}$, a \textbf{QA Agent}, $\pi_{QA}$, is applied to these contexts to produce the cycle's final output $O^{(t)}$:
\begin{equation}
\label{eq:tans}
O^{(t)} = \pi_{QA}\big(q_{init},\;\mathcal{M}_{encode}^{(t)},\;\mathcal{C}_{fuse}^{(t)} \big)
\end{equation}
Specifically, a LLM is instructed to take these latest evidence and the past background as the context, and determine whether the original query can be resolved. It either yields the \textbf{final answer} and terminates the entire reasoning loop, or signals \textbf{Failure} and continues to the next step.

\textbf{\mupdate}:
this last step in a cycle simply incorporates the newly generated memory units into the global pool, with their embedding encoded, for future retrieval and reasoning:
\begin{equation}
\label{eq:mupdate}
\mathcal{M}_{pool}^{(t)} \leftarrow \mathcal{M}_{pool}^{(t-1)} \cup \mathcal{M}_{encode}^{(t)}
\end{equation}

\paragraph{\rag}
With the above six steps from \textbf{\tret} to \textbf{\mupdate}, one cycle of the cognitive loop is realized.
For the initial step as in $t=0$, \rag starts with one round of \textbf{\tret} followed by \textbf{\tans}. If Failure is signaled, it initiates the Metacognitive loop of stateful reasoning on exploratory paths, characterized by the interlocking operations with the memory workspace, which enables to tackle complex narrative comprehension. 

In essence, our framework grasps on the principle that for long context comprehension, especially in narratives where the entire context is cohesively interconnected through the underlying plot progression \cite{xu-etal-2024-fine}, the query resolution is not a linear pipeline; rather, it is a dynamic, evolving interplay between \textbf{new evidence acquisition} and \textbf{past knowledge consolidation}, analogous to the human cognitive process. 
The overall process is further depicted in the algorithm of Appendix~\ref{sec:ComoRAG_algo}; detailed prompts used by each LLM agent are provided in Appendix~\ref{sec:prompt}.

\section{Experimental Settings}
\label{sec:exp}

\begin{table*}[!t]
\centering
\small
\begin{tabular}{cl|cccccc|cc|c}
\toprule
\textbf{Category} & \textbf{Method} & \multicolumn{2}{c}{\textbf{NarrativeQA}} & \multicolumn{2}{c}{\textbf{EN.QA}} & \textbf{EN.MC} & \textbf{DetectiveQA} & \multicolumn{2}{c}{\textbf{QA Avg.}} & \textbf{MC Avg.} \\
\cmidrule(lr){3-4} \cmidrule(lr){5-6} \cmidrule(lr){7-7} \cmidrule(lr){8-8} \cmidrule(lr){9-10} \cmidrule(lr){11-11}
& & F1 & EM & F1 & EM & ACC & ACC & F1 & EM & ACC \\
\midrule
\multirow{1}{*}{\shortstack[c]{\textbf{LLM}}}
& GPT-4o-mini          & 27.29 & 7.00  & 29.83 & 12.82 & 30.57 & 30.68 & 28.56 & 9.91  & 30.63 \\
\midrule
\multirow{3}{*}{\shortstack[c]{\textbf{Naive RAG}}}
& BGE-M3(0.3B)         & 23.16 & 15.10 & 23.71 & 16.24 & 59.82 & 54.54 & 23.44 & 15.67 & 57.18 \\
& NV-Embed-v2 (7B)     & 27.18 & \underline{17.80} & \underline{34.34} & \underline{24.57} & 61.13 & 62.50 & 30.76 & \underline{21.19} & 61.82 \\
& Qwen3-Embed-8B       & 24.19 & 15.60 & 25.79 & 17.95 & \underline{65.50} & 61.36 & 24.99 & 16.78 & 63.43 \\
\midrule
\multirow{2}{*}{\shortstack[c]{\textbf{Enhanced RAG}}}
& RAPTOR               & 27.84 & \underline{17.80} & 26.33 & 19.65 & 57.21 & 57.95 & 27.09 & 18.73 & 57.58 \\
& HippoRAGv2           & 23.12 & 15.20 & 24.45 & 17.09 & 60.26 & 56.81 & 23.79 & 16.15 & 58.54 \\
\midrule
\multirow{4}{*}{\shortstack[c]{\textbf{Multi-step}\\\textbf{RAG}}}
& Self-RAG               & 19.60 & 6.40 & 12.84 & 4.27 & 59.83 & 52.27 & 16.22 & 5.34 & 56.05 \\
& MemoRAG               & 23.29 & 15.20 & 19.40 & 11.64 & 55.89 & 51.13 & 21.35 & 13.42 & 53.51 \\
& RAPTOR+IRCoT         & \underline{31.35} & 16.00 & 32.09 & 19.36 & 63.76 & \underline{64.77} & \underline{31.72} & 17.68 & \underline{64.27} \\
& HippoRAGv2+IRCoT     & 28.98 & 13.00 & 29.27 & 18.24 & 64.19 & 62.50 & 29.13 & 15.62 & 63.35 \\
\midrule
& \textbf{ComoRAG (Ours)} & \textbf{31.43} & \textbf{18.60} & \textbf{34.52} & \textbf{25.07} & \textbf{72.93} & \textbf{68.18} & \textbf{32.98} & \textbf{21.84} & \textbf{70.56} \\
\bottomrule
\vspace{-1.5ex}
\end{tabular}
\caption{Evaluation results on four long narrative comprehension datasets. For fair comparison, all methods use GPT-4o-mini as the LLM backbone, and all non-naive RAG methods use BGE-M3 for retrieval (details in Section~\ref{sec:exp}). We highlight the \textbf{best} and \underline{second-best} results. \rag is shown consistently outperform all baselines across all datasets.}
\label{tab:main}
\end{table*}

\paragraph{Datasets}

Our experiments cover four long-context narrative understanding datasets for comprehensive evaluation, featuring both question answering through free generation (QA), and multi-choice questions by selecting the best option (MC).

\begin{itemize}[noitemsep,nolistsep,leftmargin=*]
\item \textbf{NarrativeQA} \cite{NarrativeQA}: a QA dataset consisting of books and movie scripts. For ease of computation, we follow prior works and randomly sample 500 questions from the test set, with average context length 58k tokens.
\item \textbf{EN.QA} from $\infty$BENCH \cite{inifitebench}: a QA dataset with 351 questions on classic novels, with average context length over 200k tokens.
\item \textbf{EN.MC} from $\infty$BENCH: a MC dataset with 229 questions on classic novels of similar length as EN.QA.
\item \textbf{DetectiveQA} \cite{DetectiveQA}: a MC dataset consisting of detective fiction with average length over 100k tokens. We randomly sample 20\% of all stories to reduce the computational cost.
\end{itemize}

\noindent For evaluation metrics, we report both F1 and Exact Match (EM) scores for QA datasets, and report Accuracy (ACC) for MC datasets. To ensure fairness in resolving multiple-choice questions, we only expose the options during \textbf{\tans}, such that no retrieval-related actions can utilize potential hints present in the options.

\paragraph{Baselines}

We employ four types of baselines as follows, covering different paradigms for long context QA.

\begin{itemize}[noitemsep,nolistsep,leftmargin=*]
\item \textbf{LLM}: the non-RAG setting, where the entire context (capped by length 128k) is provided to the LLM directly.

\item \textbf{Naive RAG}: the standard RAG setting that splits the raw context by chunks for retrieval. We set the max chunk length as 512 tokens in all experiments.

\item \textbf{Enhanced RAG}: RAG methods with augmented retrieval index,  including RAPTOR \cite{raptor} that constructs a semantic summary tree over text chunks, and HippoRAGv2 \cite{hipporag2} that builds the knowledge base for entities in text chunks. We also experimented with GraphRAG \cite{graphrag}; however, it requires exponential computational cost for building the retrieval index, being less practical for full evaluation. We separately report GraphRAG on a subset in Appendix~\ref{sec:impl}.

\item \textbf{Multi-step RAG}: RAG methods with multi-step or iterative retrieval strategies. IRCoT \cite{ircot} leverages Chain-of-Thought (CoT) as intermediate queries that iteratively retrieve evidence. Self-RAG \cite{selfrag} trains a dedicated critic model to control when to stop retrieval. MemoRAG \cite{memorag} trains a model that compresses the global context, which generates clues as intermediate queries.
\end{itemize}

\paragraph{Implementation Details}

For the Hierarchical Knowledge Source, we follow the procedures of HippoRAGv2 and RAPTOR respectively to build the Veridical and Semantic layers; the Episodic layer employs an adaptive sliding window for narrative summaries described in Appendix~\ref{sec:impl}.

For LLMs, our main experiments adopt \textbf{GPT-4o-mini} in all approaches to ensure fair comparison. We additionally tested GPT-4.1 and Qwen3-32B \cite{yang2025qwen3} for generalization analysis in Section~\ref{ssec:analysis}. For all RAG methods, we adopt the popular model \textbf{BGE-M3} \cite{chen-etal-2024-m3} for retrieval. Additionally, for naive RAG, we also experiment with larger but less practical embedding models, including NV-Embed-v2 \cite{lee2025nvembed} and Qwen3-Embed-8B \cite{zhang2025qwen3embeddingadvancingtext}. The LLM context length for all RAG methods, including \rag, is capped at 6k tokens.

For the Metacognitive Regulation loop, we set the framework to iterate for a maximum of 5 rounds.
More regarding implementation details are provided in Appendix~\ref{sec:impl}.

\section{Experimental Results}
\subsection{Main Results}

Evaluation results of our main experiments are shown in Table~\ref{tab:main}. Remarkably, \rag achieves the best performance upon all baselines across all datasets. Despite using the lightweight 0.3B BGE-M3 for retrieval, it significantly outperforms RAG with much larger 8B embedding models. Overall, \rag demonstrates consistent improvement for tackling long narrative comprehension, surpassing strong prior RAG methods of various paradigms.

\begin{figure}[t]
\centering
\includegraphics[width=0.6\columnwidth]{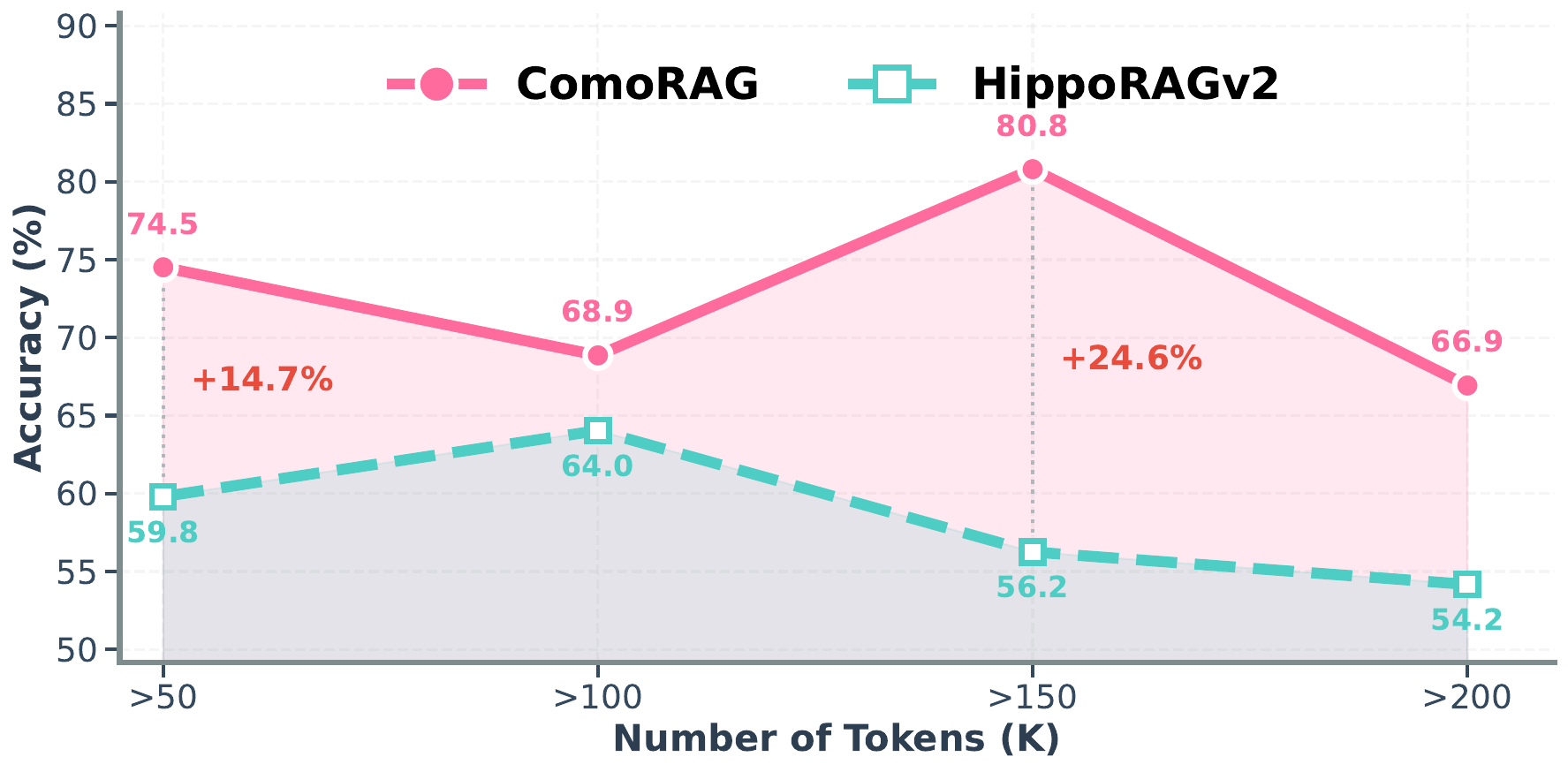}
\caption{Averaged accuracy across different document lengths on Multi-Choice datasets. \rag is shown more robust to long contexts over the baseline.}
\label{fig:token_acc_clean}
\end{figure}

Upon closer examination, \rag exhibits distinct advantages on the two $\infty$BENCH datasets featuring ultra-long contexts.
More broadly, Figure~\ref{fig:token_acc_clean} illustrates that \rag is more robust and insensitive to longer contexts, sustaining its efficacy over HippoRAGv2, with the accuracy gap peaking at +24.6\% for documents exceeding 150k tokens, which highlights the importance of stateful multi-step reasoning for query resolution over long and coherent contexts.

\subsection{Ablation Studies}
We perform ablation studies on EN.MC and EN.QA datasets by systematically removing key modules in \rag. The results are shown in Table~\ref{tab:ablation_study}.

\begin{table}[t] 
\centering
\small
\begin{tabular}{lcccc} 
\toprule 
Method & \multicolumn{1}{c}{EN.MC} & \multicolumn{2}{c}{EN.QA} \\ 
\cmidrule(lr){2-2}\cmidrule(lr){3-4} 
& ACC & F1 & EM \\
\midrule 
\textbf{ComoRAG} & \textbf{72.93} & \textbf{34.52} & \textbf{25.07} \\ 
\midrule 
\multicolumn{2}{l}{\textit{\textbf{Baselines}}} \\ 
~~HippoRAGv2 & 60.26 & 24.45 & 17.09 \\
~~RAPTOR & 57.21 & 26.33 & 19.65 \\
\multicolumn{4}{l}{\textit{\textbf{Index}}} \\ 
~~w/o Veridical & 51.97 & 22.24 & 15.88 \\ 
~~w/o Semantic & 64.63 & 30.82 & 22.65 \\ 
~~w/o Episodic & 64.63 & 31.48 & 21.47 \\ 
\multicolumn{4}{l}{\textit{\textbf{Retrieval}}} \\ 
~~w/o Metacognition & 62.01 & 26.95 & 18.53 \\ 
~~w/o Regulation & 55.02 & 27.95 & 20.59 \\
~~w/o Both & 54.15 & 25.64 & 17.35 \\
\bottomrule 
\end{tabular} 
\caption{Ablation studies of \rag.}
\label{tab:ablation_study} 
\vspace{-1.5ex}
\end{table}

\paragraph{Hierarchical Knowledge Source}
All three knowledge layers contribute supplementary enhancements to the final performance, with the Veridical Layer being the most significant retrieval index. It provides the basis for factual-grounded reasoning, as confirmed by the ~30\% relative performance drop upon its removal.

\paragraph{Metacognition}
Removing the Metacognition process essentially disables the memory workspace, where all agents operate on retrieved evidence directly, without knowledge consolidation by \emph{synthesized cues}. 
Disabling this module leads to a significant performance drop, as seen by the 22\% relative decrease in F1 score on EN.QA, and an approximate 15\% decrease in accuracy on EN.MC, underscoring the critical role of dynamic memory organization.

\paragraph{Regulation}
Removing the Regulation process cuts off the goal-oriented guidance, such that each cycle uses the same initial query for new evidence retrieval (duplicated evidence is removed), without generating probing queries that are crucial to new evidence acquisition.
Disabling this module severely impacts retrieval efficiency, causing a 24\% drop in accuracy on EN.MC and a 19\% drop in F1 score on EN.QA.

Notably, removing both Metacognition and Regulation further degrades performance, effectively reducing the system to a one-shot resolver without multi-step reasoning. Overall, the ablation study results corroborate that the enhancement offered by \rag stems from the synergy between its memory consolidation and dynamic evidence exploration, facilitated by the hierarchical knowledge index to provide enriched semantic information. Removing any of the core components would significantly weaken its narrative reasoning capabilities.

\subsection{In-Depth Analysis of Iterative Retrieval}
\label{ssec:analysis}

\begin{figure}[t]
\centering
\includegraphics[width=0.8\columnwidth]{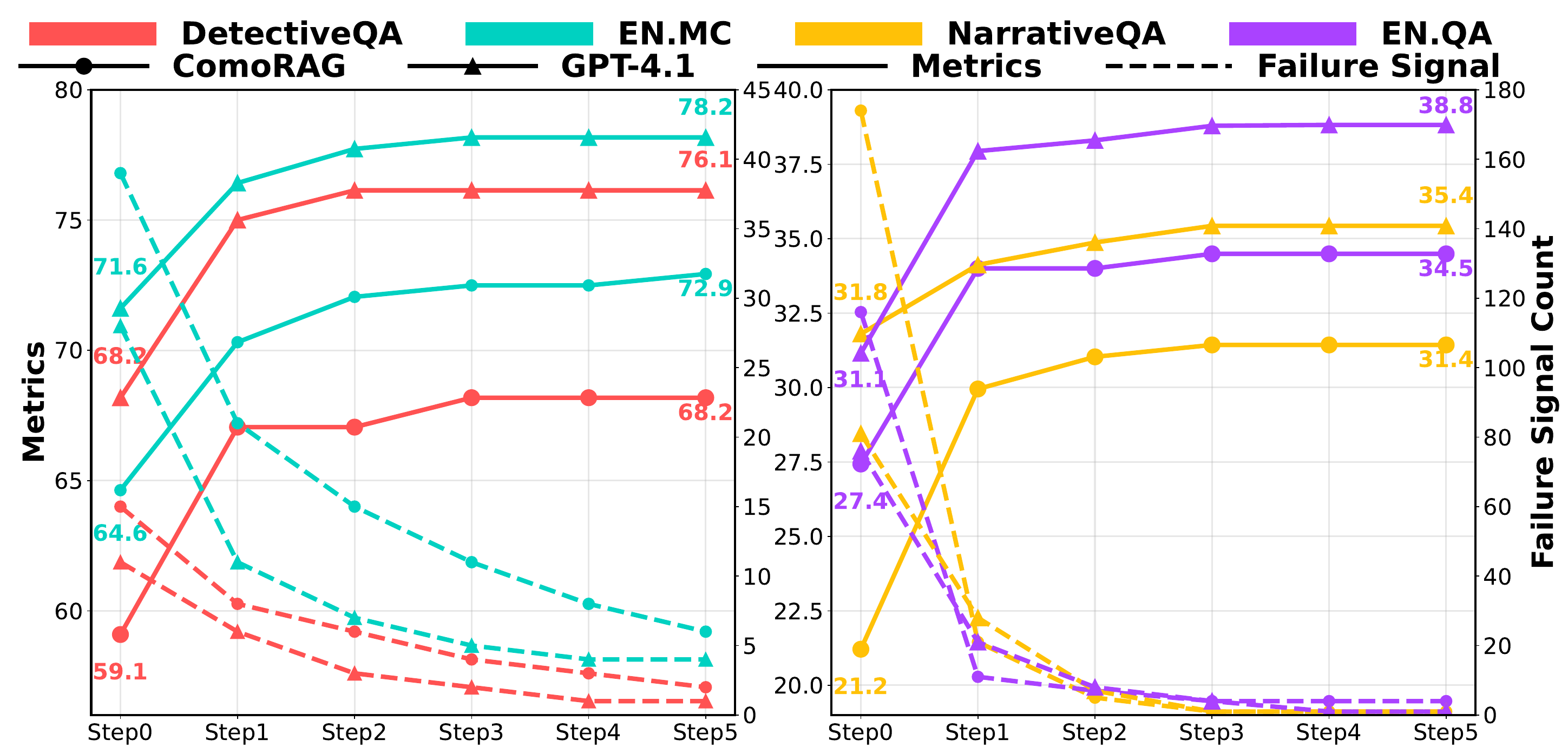} 
\caption{Performance gains from iterative probing. \emph{GPT-4.1} marks the evaluation by using the stronger GPT-4.1 as LLM agents in \rag (as opposed to GPT-4o-mini).}
\label{fig3:terative_query_rewriting}
\vspace{-1ex}
\end{figure}

To further investigate the source of \rag's effectiveness, this section presents a quantitative analysis of its core iterative retrieval process.

\paragraph{Source of Gains: From Static Bottleneck to Dynamic Reasoning}
Our analysis suggests that the stateful multi-step reasoning enabled by the Metacognitive loop is the key factor driving the observed improvement.

We first identify a ``static bottleneck'': after the initial retrieval using the original query at step 0, the single-step evaluation score shows no significant advantage over strong baselines, with less than 1\% compared to the best baseline HippoRAGv2+IRCoT.
However, upon activating the cognitive loop, there presents a sustained and significant improvement, raising the accuracy to 72.93\% on EN.MC, as shown in Figure~\ref{fig3:terative_query_rewriting}.
This further supports the findings from the ablation studies, which demonstrate a significant performance drop upon removing the entire loop. Additionally, Figure~\ref{fig3:terative_query_rewriting} illustrates that the majority of the improvement occurs within 2-3 cycles, confirming the efficiency of the process. 
The few remaining unresolved queries are tied to the inherent reasoning limitation of the base LLM, where our next analysis shows that the ceiling performance of \rag can be lifted by switching to more capable LLMs.

\paragraph{Model-agnostic Generalization}
\rag demonstrates generalization with different LLM backbones, with stronger LLMs further enhancing the reasoning process and final query resolution. To validate this, we replace GPT-4o-mini with GPT-4.1 and Qwen3-32B in the Metacognitive loop, using the same knowledge source for retrieval. The results, presented in Figure~\ref{fig3:terative_query_rewriting} and the upper section of Table~\ref{tab:compressed}, show a notable improvement particularly with GPT-4.1, boosting the F1 score on EN.QA from 34.52 to 38.82, and increases the accuracy on EN.MC from 72.93 to 78.17. These results demonstrate that \rag effectively leverages and unleashes the model's capabilities during its stateful iterative reasoning process.

\paragraph{Plug-and-Play: Flexibility}
To examine the modularity of our framework, we conduct further experiments by applying the Metacognitive loop of \rag on existing RAG methods. 
As shown in the bottom section of Table~\ref{tab:compressed}, the cognitive loop can be seamlessly integrated with different RAG index including HippoRAGv2 and RAPTOR. 
This integration consistently results in significant performance improvements across all benchmarks, with accuracy on EN.MC increasing by over 8\% for HippoRAGv2 and nearly 12\% for RAPTOR (a similar trend is observed on EN.QA). These results demonstrate that \rag could serve as a robust and flexible plug-and-play solution to enhance query resolution of existing RAG methods.

\begin{table}[!tbp]
\centering
\small
\begin{tabular}{l|cccc}
\toprule
\textbf{Method} & \textbf{NarQA} & \textbf{EN.QA} & \textbf{EN.MC} & \textbf{DetQA} \\
\cmidrule(lr){2-2} \cmidrule(lr){3-3} \cmidrule(lr){4-4} \cmidrule(lr){5-5}
& F1 & F1 & ACC & ACC \\
\midrule 
\rag & 31.43 & 34.52 & 72.93 & 68.18 \\
\textbf{ w/ Qwen3-32B} & 32.17 & 35.29 & 74.24 & 69.32 \\
\textbf{ w/ GPT-4.1} & 35.43 & 38.82 & 78.17 & 76.14 \\
\midrule
\midrule
HippoRAGv2 & 23.12  & 24.45  & 60.26 & 56.81  \\
\textbf{ + Our Loop} & \textbf{29.12} & \textbf{31.76} & \textbf{68.56} & \textbf{63.64} \\
\midrule 
RAPTOR & 27.84  & 26.33  & 57.21 & 57.95  \\
\textbf{ + Our Loop} & \textbf{30.55} & \textbf{34.31} & \textbf{69.00} & \textbf{62.50}\\
\bottomrule
\end{tabular}
\caption{Efficacy of ComoRAG on model-agnostic generalization and Plug-and-Play flexibility.}
\label{tab:compressed}
\vspace{-1.5ex}
\end{table}

\subsection{In-Depth Analysis of Query Resolution}
\label{ssec:query_type}

To deepen the understanding of narrative query resolution, we roughly categorize all questions in our experimented datasets into three query types: \textbf{factoid}, \textbf{narrative}, and \textbf{inferential}, described as follows (details in  Appendix~\ref{sec:appendix_annotation}).

\begin{itemize}[noitemsep,nolistsep,leftmargin=*]
    \item \textbf{Factoid Queries}: queries answerable by a single, specific piece of information, often knowledge-seeking, e.g., \textit{``What religion is Octavio Amber?''}
    \item \textbf{Narrative Queries}: queries that require an understanding of plot progression as a coherent background context, e.g., \textit{``Where does Trace choose to live at the end of the novel?''}
    \item \textbf{Inferential Queries}: queries demanding reasoning beyond the literal text to understand implicit motivations, e.g., \textit{``What is the main reason that Nils first visits Aiden in his apartment?''}
\end{itemize}

\begin{figure}[t]
\centering
\includegraphics[width=0.8\columnwidth]{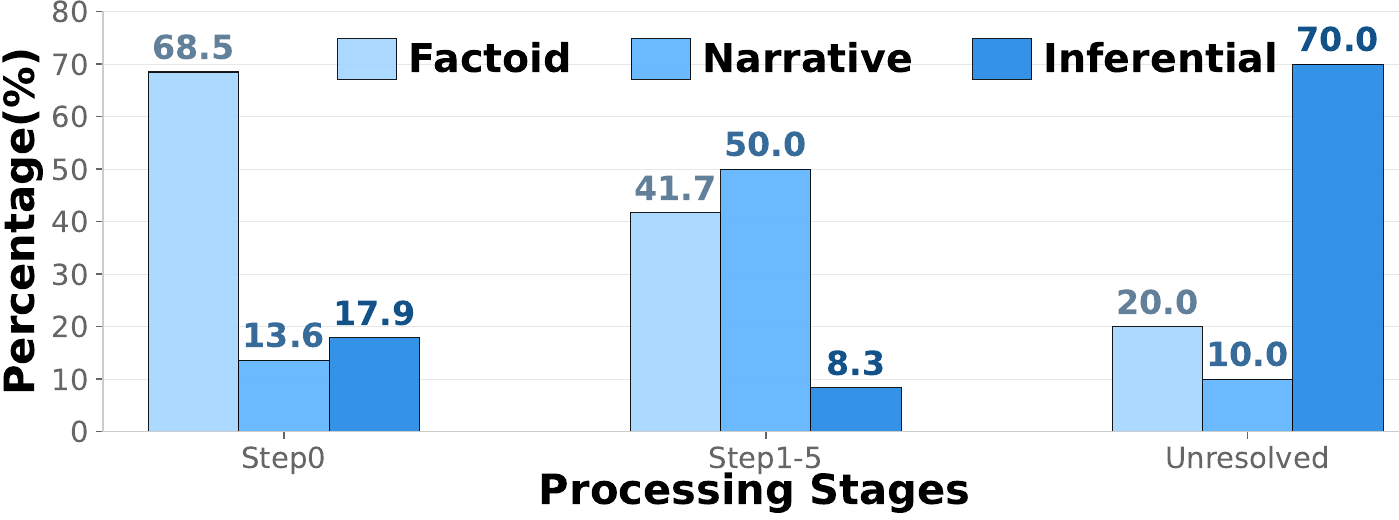}
\small
\caption{Distribution of solved query types.}
\label{fig4:Step-wise retrieval performance}
\vspace{-1ex}
\end{figure}

\noindent To systematically investigate the dynamics of \rag reasoning, we first pose the question: \textbf{what is the bottleneck in long-narrative reasoning for existing RAG methods?} Figure \ref{fig4:Step-wise retrieval performance} pictures a clear diagnosis. While one-shot retrieval suffices for factoid queries, which account for over 60\% of initial solution, our iterative cognitive loop is essential for resolving complex narrative queries involving global context comprehension and deeper reasoning. These constitute nearly 50\% of the problems that are solved exclusively through the Metacognitive loop.

\begin{figure}[t]
\centering
\includegraphics[width=0.5\columnwidth]{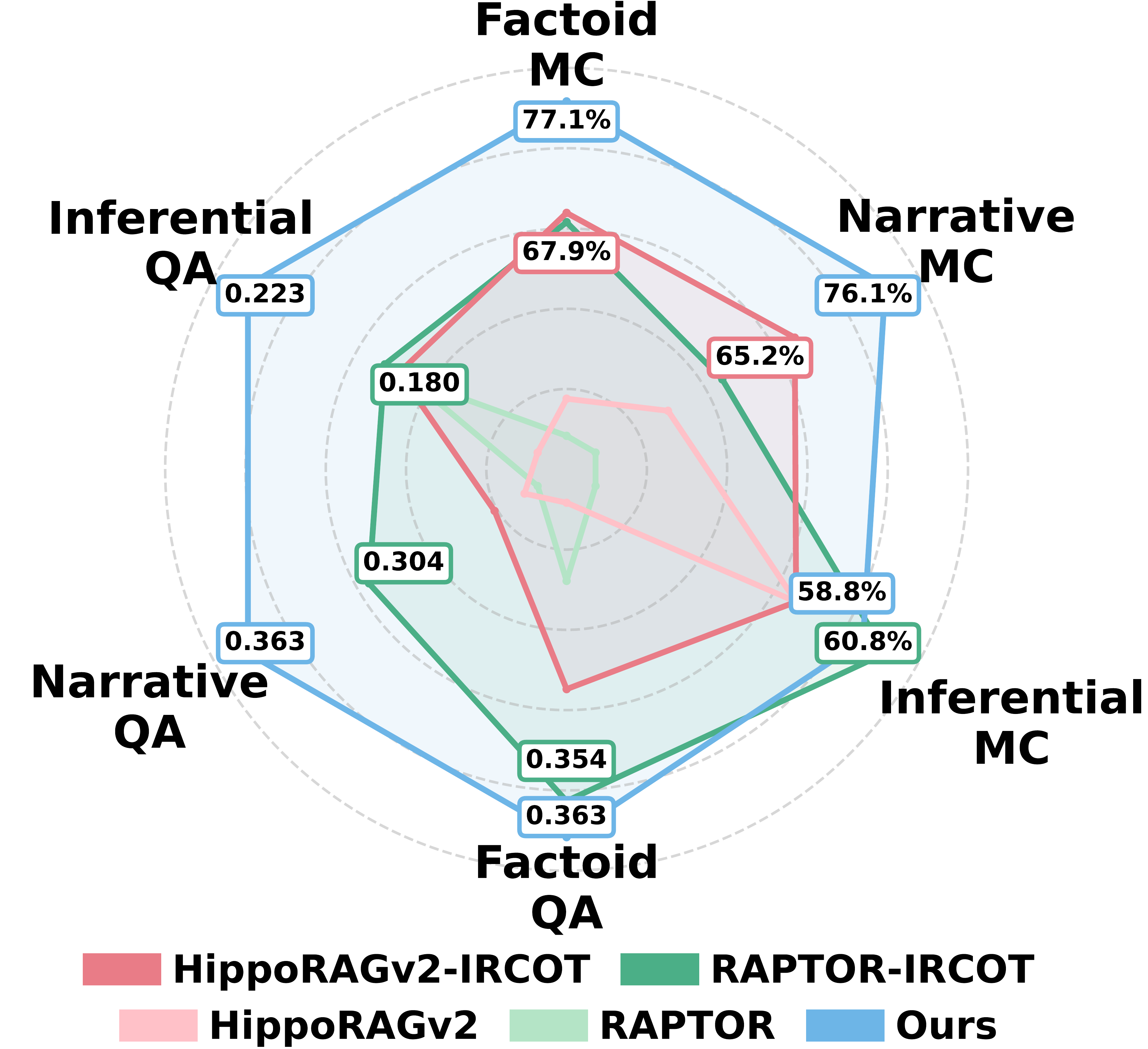}
\caption{Benchmarking RAG methods across query types.}
\label{fig5:Comparative analysis of RAG frameworks}
\vspace{-1.5ex}
\end{figure}

This leads to the second question: how does our framework's performance on this specific bottleneck compared to strong baselines? Figure \ref{fig5:Comparative analysis of RAG frameworks} demonstrates that our method’s advantage is the most pronounced precisely in this area. On narrative queries, ComoRAG substantially outperforms the strongest baselines, achieving a \textbf{19\%} relative F1 improvement on EN.QA and a \textbf{16\%} accuracy gain on EN.MC.
By addressing these queries, \rag is demonstrated a targeted and effective solution for narrative queries that have posed challenges to the conventional RAG methodology.

Qualitatively, Figure~\ref{fig:overview_comorag} illustrates the dynamic reasoning mechanism with the query $q_{init}$: \textit{``Mrs. MacIntyre never writes letters, so what is the sudden purpose of buying ink?''}
A standard, single-step retrieval would fail on this query, as it would only find a vague clue about a ``cut out newspaper'',  which is insufficient to form an answer. In contrast, \rag initiates an iterative reasoning process by dynamically probing new queries and corresponding evidence for full resolution, constructing a \textbf{complete evidence chain to deduce the final answer}: Mrs. McGinty recognized a photo, wanted to sell the story, and intended to write to the newspaper. The full reasoning details of ComoRAG on this query are further provided in Appendix~\ref{sec:case_study}.

\section{Conclusion}
In this work, we propose \rag for long narrative reasoning, aiming to address the ``stateless'' limitation of conventional RAG.
\rag is especially inspired by the human brain's Prefrontal Cortex: through a dynamic memory workspace and iterative probes, it fuses fragmented evidence into a coherent context to achieve stateful reasoning over narrative progression. Experiments validate that \rag overcomes the bottleneck of existing methods by excelling at complex narrative and inferential queries, marking a paradigm shift from information retrieval to cognitive reasoning towards deeper long context comprehension.

\bibliography{aaai2026}

\appendix
\definecolor{morandiBlueBack}{RGB}{230, 240, 255}
\definecolor{morandiBlueEdge}{RGB}{74, 144, 226}
\definecolor{morandiDivider}{RGB}{180, 200, 220}

\section{\rag Algorithm} \label{sec:ComoRAG_algo}

\begin{algorithm}[H]
\caption{\rag (Described in Section~\ref{sec:method})}
\label{alg:ComoRAG_concise}
\begin{algorithmic}[1]
\Require Initial Query $q_{init}$, Knowledge Source $\mathcal{X}$, Max Iterations $T$
\Ensure The final answer $O$ or a failure signal

\Function{ComoRAG}{$q_{init}, \mathcal{X}, T$}
    \State $\mathcal{M}_{pool}^{(0)}, \mathcal{P}_{hist}^{(0)}, \{\mathcal{C}\}^{(0)} \gets \emptyset, \emptyset, \emptyset$ \Comment{Initialize Memory Pool, Probing History, and Synthesized Cues}

    \State $\mathcal{E}^{(0)} \gets \tret(\{q_{init}\}, \mathcal{X})$ 
      \State $O^{(0)} \gets \tans(q_{init}, \mathcal{E}^{(0)})$ 
    \If{$O^{(0)} \neq \text{FailureSignal}$}
        \State \textbf{return} $O^{(0)}$ \Comment{Return immediately if successful}
    \EndIf
    
    \Statex \Comment{\textit{Triggered only if initial attempt fails}}
    \State $\mathcal{M}_{encode}^{(0)} \gets \menc(q_{init}, \mathcal{P}_{hist}^{(0)}, \mathcal{E}^{(0)})$ 
    \State $\mathcal{M}_{pool}^{(0)} \gets \mupdate(\mathcal{M}_{pool}^{(0)}, \mathcal{M}_{encode}^{(0)})$ 
    \State $\mathcal{P}_{hist}^{(0)} \gets {q}_{init} $
    \State $\{\mathcal{C}\}^{(0)} \gets \mathcal{M}_{pool}^{(0)}$
    \For{$t=1, \dots, T$} 
        \State $\mathcal{P}^{(t)} \gets \smon(q_{init}, \mathcal{P}_{hist}^{(t-1)}, \{\mathcal{C}\}^{(t-1)})$        
        \State $\mathcal{E}^{(t)} \gets \tret(\mathcal{P}^{(t)}, \mathcal{X})$
        \State $\mathcal{M}_{encode}^{(t)} \gets \menc(q_{init}, \mathcal{P}^{(t)}, \mathcal{E}^{(t)})$
        \State $\mathcal{C}_{fuse}^{(t)} \gets \mfuse(q_{init}, \mathcal{M}_{pool}^{(t-1)} \circ q_{init} \big)$
        \State $O^{(t)} \gets \tans(q_{init}, \mathcal{M}_{encode}^{(t)}, \mathcal{C}_{fuse}^{(t)})$
        
        \If{$O^{(t)} \neq \text{FailureSignal}$} \textbf{return} $O^{(t)}$ \EndIf
        
        \State $\mathcal{M}_{pool}^{(t)} \gets \mupdate(\mathcal{M}_{pool}^{(t-1)}, \mathcal{M}_{encode}^{(t)})$
        \State $\mathcal{P}_{hist}^{(t)} \gets \mathcal{P}_{hist}^{(t-1)} \cup \mathcal{P}^{(t)}$
        \State $\{\mathcal{C}\}^{(t)} \gets \mathcal{M}_{pool}^{(t)}$
    \EndFor
    \State \textbf{return} \text{FailureSignal}
\EndFunction
\end{algorithmic}
\end{algorithm}

\section{Implementation Details}
\label{sec:impl}

\subsection{Veridical Layer}
As described in Section~\ref{ssec:index}, \rag empowers Large Language Models by constructing a hierarchical knowledge source, whereby the Veridical Layer is a foundational component comprising text chunks of the original context. We largely follow the construction process of HippoRAGv2 \cite{hipporag2} to add a mapping between knowledge graphs (KGs) and text chunks to facilitate retrieval. To construct the KG, a Large Language Model (LLM) is leveraged to extract (subject-predicate-object) knowledge triples. These triples from a document are then aggregated to form a unified knowledge graph. Finally, a retrieval-optimized encoder adds supplementary edges to this graph by identifying and linking semantically similar entities (synonyms). The retrieval of the Veridical Layer thus follows HippoRAGv2 to utilize KGs towards more accurate retrieval. Statistics for this layer are detailed in Table~\ref{tab:dataset_stats}.

\begin{table}[h!] 
\small 
\centering 
\setlength{\tabcolsep}{4pt} 
\begin{tabular}{ll S[table-format=5.0] S[table-format=6.0] S[table-format=6.0] S[table-format=5.0]} \toprule \textbf{Layer} & \textbf{Count} & {\textbf{NarQA}} & {\textbf{EN.QA}} & {\textbf{EN.MC}} & {\textbf{DetQA}} \\ \midrule \multirow{3}{*}{\textbf{Veridical}} & \# of Chunks & 4446 & 26465 & 47074 & 2406 \\ & \# of Entities & 33810 & 292170 & 401040 & 30969 \\ & \# of Triples & 51012 & 372339 & 576595 & 33696 \\ \bottomrule \end{tabular} 
\caption{Statistics of the Veridical Layer across Datasets.} 
\label{tab:dataset_stats} 
\end{table}
\subsection{Episodic Layer}\label{sec:Episodic}

To construct the Episodic Layer, a sequence of text chunks is summarized. Since the context lengths can vary significantly, the choice of a sliding window size for this summarization presents a trade-off: a large window can be too coarse for short narratives, while a small window may be inefficient and fail to capture long-range dependencies in long-form content.
Therefore, we dynamically adjust the window size W according to the total number of text chunks, $N$, in the document. The specific heuristic is as follows.

\begin{itemize}
    \item For short to medium-length narratives ($N \le 200$ chunks): stepped window sizes (3, 5, 8, and 10) are used for documents up to 20, 50, 100, and 200 chunks respectively, aiming to preserve details for shorter contexts.

    \item For long narratives ($N > 200$): A logarithmic scaling function is applied to prevent the window from becoming excessively large. This sub-linear growth is intended to increase the summary scope for massive texts more slowly. The window size is calculated as follows to keep the window size between 10 to 20:
    \[
    W = \min(20, \max(10, \lfloor \log_2(N) \times 2 \rfloor))
    \]
\end{itemize}

For each window, the contained text chunks are concatenated and provided to an LLM agent (GPT-4o-mini in our experiments). The agent is instructed to generate a concise summary that maintains chronological order and identifies key events and causal relationships. The resulting summaries are then collected and sorted by their original window order to form the nodes of the Episodic Layer.

\subsection{GraphRAG Experiments}
GraphRAG is a structured-augmented RAG method similar to HippoRAGv2, which involves the construction of a comprehensive knowledge graph from source documents, which is then used to identify interconnected information for retrieval. However, its formulation requires heavy computation for building the retrieval index that includes multi-level node relations and summaries.

We conducted preliminary experiments on a data subset to evaluate its viability. The results, detailed in Table~\ref{tab:comparison_graphrag}, demonstrated that GraphRAG not only had significantly higher token consumption, but also attained lower scores compared to other baselines adopted in our experiments. Considering the trade-offs between its computational cost and performance, we ultimately did not include GraphRAG as a primary baseline for a full-scale evaluation.
\begin{table}[htbp]
    \centering
    \begin{tabular}{lrr}
        \toprule
        & \textbf{\rag} & \textbf{GraphRAG} \\
        \midrule
        \multicolumn{3}{l}{\textbf{Performance Metrics}} \\
        \addlinespace[0.3em]
        \quad F1 Score & 33.61 (100.0\%) & 14.20 (42.3\%) \\
        \quad EM Score & 21.43 (100.0\%) & 8.00 (37.3\%) \\
        \midrule
        \multicolumn{3}{l}{\textbf{Token Usage}} \\
        \addlinespace[0.3em]
        \quad Tokens  & 5.90M (100.0\%) & 27.12M (459.7\%) \\
        \midrule
        \multicolumn{3}{l}{\textbf{Average Time Taken (sec)}} \\
        \addlinespace[0.3em]
        \quad Index & 291 (100.0\%) & 1936 (665.3\%) \\
        \quad Retrieve & 25 (100.0\%) & 29 (116.0\%) \\
        \bottomrule
    \end{tabular}
    \caption{Comparison of Performance, Token Usage, and Average Time for \rag and GraphRAG.}
    \label{tab:comparison_graphrag}
\end{table}

\subsection{Hyperparameters for \rag}
The key hyperparameters for our \rag framework are detailed in Table 6. All cognitive agents employ GPT-4o-mini, with retrieval powered by the widely-used BGE-M3 embedding model. 
For retrieval settings, The dynamic cognitive loop is configured to run for a maximum of 5 iterations, generating up to 3 new probing queries per cycle.
The context for QA is capped at 6k tokens, in consistent with all RAG baselines in our experiments. This context is assembled via a proportional 8:2:2:1 allocation of evidence from the Veridical, Semantic, Episodic, and fused Historical memory, respectively. The ``Mem-Fuse Threshold'' is set to 0.5, indicating the proportion of evidences retrieved from the memory pool that are forwarded to the Integration Agent for memory fusion and summary generation.
\begin{table}[H]
    \centering
    \small 
    
    \begin{tabularx}{\columnwidth}{@{} l X @{}} 
        \toprule
        \textbf{Hyperparameter} & \textbf{Value} \\
        \midrule
        LLM Agents ($\pi_{probe}$, etc.) & GPT-4o-mini \\
        Retrieval Model & BGE-M3 \\
        Chunk Size & 512 tokens \\
        Context Length & 6,000 tokens \\
        Random Seed & 0 \\
        \midrule
        Max Iterations & 5 \\
        Max Probing Queries & 3 \\
        Context Construction & Proportional Allocation (8:2:2:1 ratio for V:S:E:H) \\
        \mfuse Threshold & 0.5 \\
        \bottomrule
    \end{tabularx}
        \caption{Hyperparameter settings for \rag in our experiments. V, S, E, H refer to Veridical, Semantic, Episodic, and Historical evidence.}
    \label{tab:hyperparameters_option2}
\end{table}

\section{Query Types for Narratives}
\label{sec:appendix_annotation}

\begin{table}[h!]
\centering
\begin{tabular}{l|ccc|c}
\toprule
\textbf{Dataset} & \textbf{Factoid} & \textbf{Narrative} & \textbf{Inferential} & \textbf{Total} \\
\midrule
EN.QA & 224 & 84 & 43 & 351 \\
EN.MC & 132 & 46 & 51 & 229 \\
\bottomrule
\end{tabular}
\caption{Distribution of query types across the two datasets.}
\label{tab:annotation_distribution}
\end{table}

To facilitate a fine-grained analysis of our model's performance, we (authors of this work) manually annotated the types of all questions in the EN.QA and EN.MC datasets. 
Each question is classified into one of the three categories based on the cognitive processes required to answer it, described in Section~\ref{ssec:query_type}:
\begin{itemize}
    \item \textbf{Factoid}: questions answerable by locating a single, specific piece of information from the text.
    \item \textbf{Narrative}: questions that demand an understanding of plot progression, requiring the aggregation of information from multiple text parts.
    \item \textbf{Inferential}: questions that necessitate reasoning beyond the literal text to understand implicit motivations or causal links.
\end{itemize}

The final distribution of the annotated query types is presented in Table~\ref{tab:annotation_distribution}.

\onecolumn
\section{Prompting Templates}\label{sec:prompt}

\begin{center}
\begin{tcolorbox}[breakable,colback=morandiBlueBack,
    colframe=morandiBlueEdge, title=\textbf{\smonw} Instruction Template for Probing Query Generation in Regulation Agent, sharp corners=south]

\textbf{Role:} \\
You are an expert in multi-turn retrieval-oriented probe generation. Your job is to extract diverse and complementary retrieval probes from queries to broaden and enrich subsequent corpus search results.

\vspace{0.8em}
\textbf{Input Materials:}
\begin{itemize}
    \item \textbf{Original Query}: A question or information need that requires comprehensive information retrieval.
    \item \textbf{Context}: Available background information, partial content, or relevant summaries.
    \item \textbf{Previous Probes}: Previously generated probes from earlier iterations (if any).
\end{itemize}

\vspace{0.8em}
\textbf{Task:} \\
Based on the query and context, generate \textbf{up to 3 non-overlapping retrieval probes} that explore the query from distinct angles.

\textbf{Critical Requirements:}
\begin{itemize}
    \item \textbf{Semantic Differentiation}: Ensure new probes are semantically distinct from any previous probes provided.
    \item \textbf{Complementary Coverage}: New probes should cover different information dimensions not addressed by previous probes.
    \item \textbf{Relevance Maintenance}: All probes must remain directly relevant to answering the original query.
\end{itemize}

\textbf{Each probe should:}
\begin{itemize}
    \item Target different information dimensions relevant to the query type:
    \begin{itemize}
        \item \emph{Character-related}: actions, motivations, relationships, timeline, consequences
        \item \emph{Event-related}: participants, causes, sequence, location, outcomes
        \item \emph{Object-related}: description, origin, usage, significance, connections
        \item \emph{Location-related}: events occurred, people involved, time periods, significance
    \end{itemize}
    \item Expand search scope beyond obvious keywords to capture related content.
    \item Avoid semantic overlap with previous probes while maintaining query relevance.
    \item Be formulated as effective search terms or phrases.
\end{itemize}

\vspace{0.8em}
\textbf{Probe Generation Strategy:}
\begin{itemize}
    \item \textbf{When previous probes exist:}
    \begin{enumerate}
        \item Analyze Previous Coverage: Identify what semantic domains/angles have been covered.
        \item Gap Identification: Find unexplored but relevant information dimensions.
        \item Alternative Angles: Generate probes from different conceptual perspectives.
        \item Semantic Distance: Ensure sufficient semantic distance from previous probes.
    \end{enumerate}
    \item \textbf{When no previous probes exist:}
    \begin{itemize}
        \item Probe 1: Direct elements explicitly mentioned in the query.
        \item Probe 2: Contextual elements that might contain the answer.
        \item Probe 3: Related concepts or alternative formulations.
    \end{itemize}
\end{itemize}

\vspace{0.8em}
\textbf{Output Format:}
\begin{verbatim}
```json
{
    "probe1": "Content of probe 1",
    ...
}
```
\end{verbatim}

\vspace{0.5em}
\textbf{Notes:}
\begin{itemize}
    \item For simple queries, you may generate only 1--2 probes.
    \item If previous probes have covered most relevant angles, generate fewer new probes to avoid redundancy.
    \item Prioritize quality and semantic distinctiveness over quantity.
\end{itemize}
\end{tcolorbox}
\end{center}

\begin{center}
\begin{tcolorbox}[breakable,
    colback=morandiBlueBack,
    colframe=morandiBlueEdge,
    title=\textbf{\mencw} Instruction Template for Synthesized Cue Generation in Comprehension Agent,
    sharp corners=south,]
    
\textbf{Role} \\
You are an expert narrative analyst capable of identifying, extracting, and analyzing key information from narrative texts to provide accurate and targeted answers to specific questions.

\vspace{0.8em}
\textbf{Material} \\
You are given the following:
\begin{enumerate}[label=\arabic*.]
    \item A final objective to be resolved
    \item A specific question that needs to be answered
    \item Content: Direct excerpts, facts, and specific information from the narrative text
\end{enumerate}

\vspace{0.8em}
\textbf{Task}
\begin{enumerate}[label=\arabic*.]
    \item Carefully analyze the question to identify:
    \begin{itemize}
        \item What type of information is being asked (character actions, locations, objects, events, motivations, etc.)
        \item Which narrative elements are relevant to answering it
        \item The specific details that need to be extracted
    \end{itemize}
    
    \item Systematically scan the content for:
    \begin{itemize}
        \item Direct mentions of relevant elements (names, places, objects, events)
        \item Contextual probes that help answer the question
        \item Temporal and spatial relationships
        \item Cause-and-effect connections
    \end{itemize}
    
    \item Analyze the identified information considering:
    \begin{itemize}
        \item Explicit statements (directly stated facts)
        \item Implicit information (suggested through context, dialogue, or narrative)
        \item Logical connections between different narrative elements
        \item Chronological sequence of events if relevant
    \end{itemize}
    
    \item Synthesize findings to construct a precise answer to the question.
\end{enumerate}

\vspace{0.8em}
\textbf{Response Format} \\
Provide a structured analysis with up to 5 key findings:\\
\texttt{```}\\
\hspace{4em} \textit{Key Finding}: \textit{\textless Most directly relevant information answering the question\textgreater}\\[0.5em]
\hspace{2em} \textit{Key Finding}: \textit{\textless Supporting evidence or context\textgreater}\\[0.5em]
\hspace{2em} \textit{Key Finding}: \textit{\textless Additional relevant details\textgreater}\\[0.5em]
\hspace{2em} \textit{Key Finding}: \textit{\textless Clarifying information if needed\textgreater}\\[0.5em]
\hspace{2em} \textit{Key Finding}: \textit{\textless Resolution of any ambiguities\textgreater}\\
\texttt{```}\\

\end{tcolorbox}
\end{center}

\begin{center}
\begin{tcolorbox}[breakable,colback=morandiBlueBack,
    colframe=morandiBlueEdge, title=\textbf{\mfusew} Instruction Template for Cue Generation in Integration Agent, sharp corners=south]
\textbf{Role:} \\
You are an expert narrative synthesis specialist who excels at integrating and analyzing information from multiple narrative sources to create coherent and comprehensive insights.

\vspace{0.8em}
\textbf{Input Material:}
\begin{itemize}
    \item \textbf{Previous Analysis:} Results from earlier memory fusion operations that contain analyzed narrative information.
    \item \textbf{Current Query:} A question or information request that needs to be addressed.
\end{itemize}

\vspace{0.8em}
\textbf{Task:}
\begin{enumerate}
    \item \textbf{Review and understand the previous memory fusion outputs:}
    \begin{itemize}
        \item Identify key narrative elements and their relationships.
        \item Note any established facts, character developments, or plot points.
        \item Recognize patterns and connections across different analyses.
    \end{itemize}

    \item \textbf{Analyze the current query in context:}
    \begin{itemize}
        \item Determine how it relates to previously established information.
        \item Identify any new aspects or angles that need to be addressed.
        \item Consider how previous insights can inform the current response.
    \end{itemize}

    \item \textbf{Synthesize the information:}
    \begin{itemize}
        \item Integrate relevant previous findings with new analysis.
        \item Create a coherent narrative that addresses the current query.
        \item Ensure continuity and consistency with previous analyses.
        \item Highlight any new insights or developments.
    \end{itemize}

    \item \textbf{Provide a comprehensive response that:}
    \begin{itemize}
        \item Directly answers the current query.
        \item Incorporates relevant previous context.
        \item Maintains narrative coherence.
        \item Offers clear and insightful analysis.
    \end{itemize}
\end{enumerate}

\vspace{0.8em}
    \textbf{Response Format:} \\
Provide a cohesive narrative response that integrates previous insights with new analysis to address the current query. Focus on creating a flowing, well-structured response.
\end{tcolorbox}
\end{center}

\begin{center}
\begin{tcolorbox}[breakable,colback=morandiBlueBack,
    colframe=morandiBlueEdge, title=\textbf{\tansw} Prompt Template for Query Resolution in QA Agent, sharp corners=south]
\textbf{Role:} \\
You are an expert on reading and understanding books and articles.

\vspace{0.8em}
\textbf{Task:} \\
Given the following detailed article, semantic summary, Episodic summary from a book, and a related question with different options, you need to analyze which option is the best answer for the question.

\vspace{0.8em}
\textbf{Inputs:}
\begin{itemize}
    \item \textbf{Detail Article:} \texttt{\{context\}}
    \item \textbf{Summary by Semantic:} \texttt{\{semantic\_summary\}}
    \item \textbf{Summary by Episodic:} \texttt{\{Episodic\_summary\}}
    \item \textbf{History Info:} \texttt{\{history\_info\}}
    \item \textbf{Question:} \texttt{\{question\}}
\end{itemize}

\vspace{0.6em}
\textbf{Limits:}
\begin{itemize}
    \item Do not infer. Respond only based on the provided content strictly.
    \item Pick the choice only if you find at least 2 places that support the answer.
\end{itemize}
\textbf{Response Format:}
\begin{enumerate}
    \item \textbf{Content Understanding}: \\
    Start with a brief summary of the content in no more than three sentences. Begin this section with \texttt{\#\#\# Content Understanding}.
    
    \item \textbf{Question Analysis}: \\
    Based on the question, analyze and list all relevant items using a markdown list. Begin this section with \texttt{\#\#\# Question Analyse}.
    
    \item \textbf{Options Analysis}: \\
    Extract the key points related to 4 options, also using a markdown list. Begin this section with \texttt{\#\#\# Options Analyse}. \\
    \textbf{Note:} Only analyze based on the provided materials, do not make guesses.
    
    \item \textbf{Final Answer}: \\
    Provide your final answer with a heading. Begin this section with \texttt{\#\#\# Final Answer}, followed by the best option in the format of \texttt{[A]} or \texttt{[B]} or \texttt{[C]} or \texttt{[D]}. If you cannot answer, give a failure signal: \texttt{*}.
\end{enumerate}

\end{tcolorbox}
\end{center}
\newpage

\section{Case Study on Narrative Reasoning}\label{sec:case_study}

\begin{table*}[!htbp]
\centering
\small
\begin{tabular}{>{\raggedright\arraybackslash}p{0.98\textwidth}}
\toprule
\multicolumn{1}{c}{\textbf{Input Data (No Options)}} \\
\textbf{Query:} Mrs. MacIntyre never writes letters, so what is the sudden purpose of buying ink? \\
\textbf{Options:}
[A] Response to the government's letter
[B] Sending a birthday card to my niece
[C] Write a letter to the Sunday Comet newspaper.
[D] Blurring Photos
\\
\midrule
\multicolumn{1}{c}{\textbf{\rag's Choice Result}} \\
\midrule
\textbf{Memory Pool $M_{pool}^{(0)}$:} \\
- A page of the newspaper \sethlcolor{lightgreen}\hl{had a portion cut out}...\\
\midrule
\fbox{\textbf{Step1}}\\
\textbf{Probes $P^{(1)}$:} \\
- What did Mrs. McGinty recognize in the clipping, and how did she try to profit from it?\\
...\\
\textbf{Retrieved Passages:} \\
...The narrative offers insight into Miss Pamela Hosford’s role at the Sunday Comet, as she casually recalls receiving a vague letter from a woman about a photograph but fails to retrieve it...\\
\textbf{Cues $C^{(1)}$:} \\
\sethlcolor{lightblue}\hl{- Key Finding:Mrs. McGinty usually had Joe help her reply to letters.};\\
\sethlcolor{lightblue}\hl{- Key Finding:Mrs. McGinty may have sent a letter to the newspaper.};... \\
\midrule
\textbf{Memory Pool $M_{pool}^{(1)}$:} \\
- A page of the newspaper \sethlcolor{lightgreen}\hl{had a portion cut out}...\\
- Mrs. MacIntyre sent a letter to \sethlcolor{lightgreen}\hl{the Sunday Comet}...\\
\midrule
\fbox{\textbf{Step2}}\\

\textbf{Probes $P^{(2)}$:} \\
- What did Miss Pamela Hosford recall about a letter sent to the Sunday Comet, and what might it imply about Mrs. McGinty?\\
...\\
\textbf{Retrieved assages:} \\
...Miss Pamela Hosford’s role at the Sunday Comet, as she casually recalls \sethlcolor{lightblue}\hl{receiving a vague letter from a woman about a photograph} but fails to retrieve it...She liked eavesdropping, knew who was in the photo, and \sethlcolor{lightblue}\hl{asked what we‘d pay}...\\
\textbf{Cues $C^{(2)}$:} \\
\sethlcolor{lightblue}\hl{- Key Finding:Mrs. McGinty wanted to make some money from the photo.};...\\
\textbf{Chosen:} \textcolor{green}{\textbf{C.}(Correct)}
\\
(C) Write a letter to the Sunday Comet newspaper: Strong textual probes support this option. Mrs. McGinty cut out a part of the newspaper, recognized someone in a photo, asked about payment, and unusually bought ink—suggesting she intended to write to the paper.
\captionsetup{font=small,labelfont=bf}
\textcolor{green}{Final Answer: [C]}
\\
\bottomrule
\end{tabular}
\caption{Case Study on Narrative Reasoning. We present a case to demonstrate our model's performance in long-context understanding, showing the final round of the Metacognitive Control Loop. Different colors are used to highlight the nature of the processed information: \sethlcolor{lightblue}\hl{Blue} is used for the key evidence that contributes to the correct answer, while \sethlcolor{lightgreen}\hl{Orange} is used for the key cues.}
\label{tab:case_study}
\end{table*}

\end{document}